# Artificial Intelligence for COVID-19 Detection-A state-of-the-art review

Parsa Sarosh[1], Shabir A. Parah[1], Romany F. Mansur[2], G. M. Bhat[3].
Department of Electronics and Instrumentation Technology, University of Kashmir, JK, India-190006
[2]Faculty of Science, New Valley University, El-Kharja, 72511, Egypt
[3]Department of Electronics Engineering, Institute of Technology, University of Kashmir, JK, India-190006

## Abstract

The emergence of COVID-19 has necessitated many efforts by the scientific community for its proper management. An urgent clinical reaction is required in the face of the unending devastation being caused by the pandemic. These efforts include technological innovations for improvement in screening, treatment, vaccine development, contact tracing and, survival prediction. The use of Deep Learning (DL) and Artificial Intelligence (AI) can be sought in all of the above-mentioned spheres. This paper aims to review the role of Deep Learning and Artificial intelligence in various aspects of the overall COVID-19 management and particularly for COVID-19 detection and classification. The DL models are developed to analyze clinical modalities like CT scans and X-Ray images of patients and predict their pathological condition. A DL model aims to detect the COVID-19 pneumonia, classify and distinguish between COVID-19, Community-Acquired Pneumonia (CAP), Viral and Bacterial pneumonia, and normal conditions. Furthermore, sophisticated models can be built to segment the affected area in the lungs and quantify the infection volume for a better understanding of the extent of damage. Many models have been developed either independently or with the help of pre-trained models like VGG19, ResNet50, and AlexNet leveraging the concept of transfer learning. Apart from model development, data preprocessing and augmentation are also performed to cope with the challenge of insufficient data samples often encountered in medical applications. It can be evaluated that DL and AI can be effectively implemented to withstand the challenges posed by the global emergency.

*Keywords*: COVID-19, RT-PCR, Artificial Intelligence, Deep Learning, CNN, Classification.

## 1. Introduction

The preliminary cases of COVID-19 pneumonia were identified in China (Wuhan District) in December 2019 [1, 2]. In January 2020, the World Health Organisation (WHO) declared COVID-19 as a public health emergency because of numerous cases being spread around the world [3]. As per a report of the WHO, more than 200 countries have been affected by the COVID-19 pandemic [4, 5]. Severe Acute Respiratory Syndrome Coronavirus 2 (SARS-COV-2) virus that causes the COVID-19 infectious disease is a single-stranded RNA virus. The total number of cases as of 25 November 2020 GMT 13:03 is 60,108, 193 with deaths being equal to 1,414,925 [6]. Despite restrictions in all spheres of life including emphasis on cleanliness and social distancing, still the number of cases is increasing rapidly. The impact is more on elder people with co-morbidities like Diabetes, coronary atherosclerotic heart disease, liver disease, lung disease, kidney disease, nervous system disease, and blood pressure, etc. When there is exposure to the virus, the human body, as a means of defense produces antibodies [7]. To identify the existence of the SARS-COV-2 virus inside the human body either the gene sequence of the virus or the presence of antibodies is identified. Therefore, the diagnosis of coronavirus is done using the method of gene sequencing or the common Reverse Transcript Polymerase Chain Reaction (RT-PCR) or antibody test [8]. The RT-PCR test is manual and time-consuming with the requirement of many resources like Personal Protective Equipment (PPE), testing swabs, and sophisticated laboratory equipment. Further issues with conventional manual testing include a high false-negative rate, more time consumption in sample collection and transportation [9, 10]. Other diagnostic tools include the investigation of symptoms and contact tracing. Although some cases may be asymptomatic, the manifestations of the COVID-19 infection can comprise of a wide range of symptoms [11]. These include cough, headache, myalgia, fever, dyspnea, difficulty in breathing, pneumonia, and some extreme cases can also lead to acute respiratory distress syndrome (ARDS), multi-organ failure, and death [12]. Despite global efforts of devising a cure for COVID-19, there is, as of now, no particular treatment and immunization. [13, 14]. It is therefore extremely important to identify and isolate the person who might have the coronavirus to halt further spread. According to the World Health Organisation (WHO), the best method to diagnose pneumonia is through the radiological examination using Chest X-rays (CXR) and CT images [15]. Radiological imaging can be used to identify the effect of COVID-19 on the epithelial cells that line the respiratory tract of the lungs of patients. These include ground-glass opacities in the lower projections in early stages and pulmonary consolidation towards the later stages of COVID-19 pneumonia [16, 17].

The rest of the paper is prepared as follows. Section 2 describes the application areas of DL and AI to tackle the COVID-19 pandemic. Section 3 describes the application of Deep Learning and AI for Covid-19 Detection and Screening. Section 4 presents the role of AI for COVID-19 detection Using CT scans. Section 5 describes the systems for COVID-19 detection, classification, segmentation, and Quantification using X-Ray images. The conclusion is presented in Section 6.

## 2. Application Areas of DL and AI for COVID-19 management:

As the number of infections increases, the demand for healthcare services also grows. The healthcare systems of even the most developed countries are not able to withstand the challenge posed by the COVID-19 pandemic [18]. These overburdened systems often struggle with a shortage of ventilators, hospital beds, Personal Protective Equipment (PPEs) for healthcare professionals, and testing kits [19]. Because of the immense number of cases, the role of human intervention in the process of healthcare delivery should be minimized as much as possible. The fight against the COVID-19 pandemic can be undertaken with the help of Artificial Intelligence (AI) particularly the Deep Learning (DL)-based models [20, 21]. The AI can be supplemented by other technologies like the Internet of Things (IoT), 6G-based mobile technologies, and smart data analytics to develop a smart healthcare system. These prominent areas where the help of AI can be sought include Screening, diagnosis, treatment, forecasting, contact tracing, survival prediction, quantification of infection, and in the clinical trials for the development of vaccines as shown in fig 1 [22, 23]. In terms of the detection and diagnosis of Coronavirus, we can make use of the fully automatic systems which employ the promising DL-based models [24]. The AI models offer advantages like fast and efficient processing of big-data which is critical in a pandemic where the number of cases rises exponentially [25]. The other advantage that AI offers is reliable performance and comparable to or beyond human level-accuracy in many tasks.

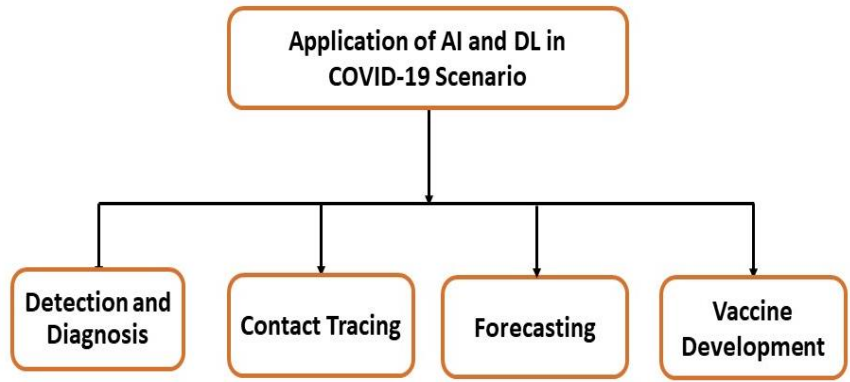

Fig 1: Application of AI and DL in COVID-19 scenario

In this paper, we present different AI-based models which help in the various aspects of the overall COVID-19 management. We further illustrate in detail the different models developed for Classification, detection, and delineation of the COVID-19 infection using Xray and CT images and the performance evaluation of the different models.

## 2.1 Detection and Diagnosis:

The developed models can be distinguished based on the modality utilized by the DL-framework [26]. The DL models can be broadly classified as employing either the X-Ray image or the CT scan for COVID-19 diagnosis. Furthermore, the developed models are either custom made for COVID-19 detection or utilize pre-trained models as backbone as shown in fig 2. Some researchers have used off-the-shelf models pre-trained on large image databases like ImageNet dataset and repurposed them for COVID-19 detection and classification [27]. The DL system can be an off-the-shelf pre-trained model used directly like the AlexNet, Visual Geometry Group (VGG-16/19), U-Net, ResNet, Xception, and GoogLeNet [28]. The pre-trained models can be further trained using smaller datasets often encountered in medical applications with the help of deep transfer learning. Other customized CNN models have also been evaluated which include the WOA-CNN, CRNet, DeCoVNet, COVID-Net, CoroNet among others [29]. The Xray images are simple to develop because of the availability of the prevalent necessary setup in almost all medical centers and are less expensive for the patient. However, the detection of COVID-19 from Xray images is difficult to perform because of subtle visual indicators. On the other hand, 3D-CT scans are developed using sophisticated machinery and are more expensive to generate, but they can provide a detailed 3D-pulmonary view. Furthermore, the sensitivity of CT images is much higher than that of RT-PCR testing

[30]. According to a study conducted in Wuhan, the sensitivity of RT-PCR for the COVID-19 infection rate was 71% and CT images were about 98%.

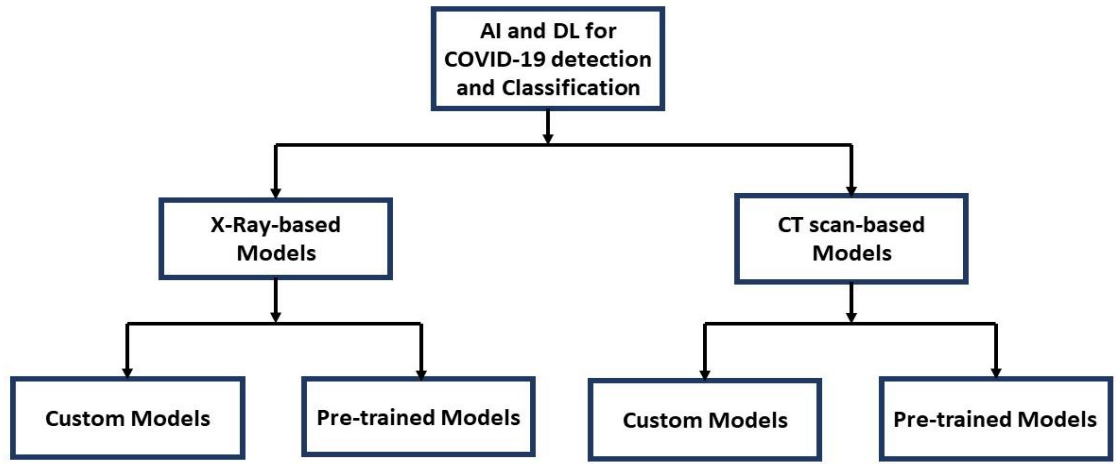

Fig 2: DL Models utilized for COVID-19 detection

The high false-negative rate of RT-PCR tests indicate less efficiency in identifying the COVID-19 affected patient. This leads to delay in the administration of proper treatment to the patient and can also lead to more spreading of the virus because of a lack of proper identification. Therefore, it is concluded that the analysis of medical images like X-rays and CT scans can provide a more accurate alternative diagnostic tool for COVID-19 pneumonia [31, 32]. The clinical findings from CT and Xray images include bilateral and peripheral ground-glass opacities (GGO), consolidation and crazy-paving pattern. However, manual delineation of the affected areas in the lungs by radiologists can be a very hectic and time-consuming process and particularly for the enormous number of patients arising from this pandemic [33]. The DL architectures can extract high-level features from both modalities and can be used to detect and diagnose COVID-19 pneumonia. The AI-based system can assist radiologists and help provide decision support to the clinicians in this pandemic for patient triage and treatment [34].

## 2.2 Contact Tracing:

According to the WHO guidelines, the spread of the virus can be contained through precautions like frequent cleansing of hands, use of personal protective equipment like gloves and masks, and Social distancing [35]. These protective measures can save others who come in contact with a person having the virus and can stop its further spread. The virus is mainly transmitted via discharges through the mouth and nose during coughing or sneezing [36]. Proper identification of the person who himself is infected and the ones he has interacted with can restrict the further spread of the SARS-Cov-2 virus. It is an important step in this regard that digital contact tracing should be implemented such that people who have come in contact with the virus can be identified and further spread of the virus can be contained [37]. Along with AI, DL, and ML, the use of various integrated technologies like IoT and communication systems like Bluetooth and GPS can be sought for this purpose [38, 39]. Many countries have created platforms for digital contact tracing which are illustrated in table 1.

Table 1: Digital Contact Tracing Applications (App) [12]

| Country | Contact Tracing App | Technology |
|---|---|---|
| UK | NHS Covid-19 App | Bluetooth |
| Australia | COVIDSafe | Bluetooth |
| India | Aarogya Setu App | Bluetooth |
| Germany | CoronaApp | Bluetooth, Google |
| China | Conjunction with Alipay | GPS, GSM |
| Italy | Immuni | Bluetooth, Google |
| Saudi Arabia | Corona Map | Bluetooth |

## 2.3 Forecasting

Forecasting models are developed to predict the growth of COVID-19 cases that might arise soon. This information will help in preparing the healthcare organizations and government institutions to generate and allocate resources like hospital beds, PPEs for healthcare workers, Ventilators, and Intensive care units to critical patients before the actual scenario arises [40, 41]. **Vinay Kumar Reddy Chimmula** et al. [42] present a DL-based approach to forecast the future COVID-19 cases in Canada using the datasets provided by John Hopkins University and Canadian health authorities. The forecasting method is based on a Long Short-term Memory (LSTM) network. Their transmission model predicted that the outbreak in Canada will have an ending point around June 2020. They conclude that the rate of transmission in Canada is taking a linear trend contrary to the exponential trend seen in the USA. **Domenico Benvenuto** et al. [43] propose an econometric model to predict the spread of the virus. They utilize the ARIMA model for prediction of epidemiological trend of occurrence of cases with the help of datasets provided by John Hopkins University. The ARIMA (1,0,4) was chosen to be the best model to calculate the prevalence of COVID-19 and ARIMA (1,0,3) has the best performance for calculating the incidence of COVID-19. **Li Yan** et al. [44] have proposed an XGBoost ML-based model that can select biomarkers that can predict the mortality of individuals almost ten days in advance with more than 90% accuracy. These biomarkers include lymphocyte, lactic dehydrogenase (LDH), and high-sensitivity C-reactive protein (hs-CRP). They evaluate that high values of LDH acts as an indicator for the identification of patients requiring immediate medical attention. **Tanujit Chakraborty** et al. [45] present a data-driven method to forecast the future cases that might arise. The data analysis has been done for various countries like France, Canada, India, South Korea, and the UK and also include risk assessment in terms of fatality rate. The method is based on a moving average model and wavelet-based forecasting model for making predictions (10 days prior) about the total number of confirmed cases. This will help prepare and allocate necessary services and equipment and can also be an early warning for the people, healthcare professionals, and government organizations. **Matheus Henrique Dal Molin Ribeiro** et al. [46] evaluate several models for time-series forecasting with 1-3 and 6 days ahead. The evaluated models include Autoregressive Integrated Moving Average (ARIMA), Random Forest (RF), ridge regression (RIDGE), Cubist regression (CUBIST), Support-Vector Regression (SVR), and stacking-ensemble learning. The best performance is offered by the SVR model and worst by the RF models in terms of evaluation parameters like improvement index, and mean absolute error among others.

**2.4 Vaccine Development**:

Yet another area where AI and DL can help in assisting healthcare professionals is to identify the use of existing drugs for the treatment of COVID-19 or develop new vaccines [47]. **Yi-Yu Ke** et al. [48] have developed an AI-based method to repurpose the potential older drugs for COVID-19 treatment. The older drugs with anti-coronavirus behavior were identified by the use of an AI platform and two learning databases. One of the databases had compounds with properties against SARS-CoV, SARS-CoV2, Influenza virus, and the other database contained 3C-like protease inhibitors. The AI-assisted method can rapidly identify which older drugs can be repurposed for the treatment of COVID-19. They isolate 8 drugs including vismodegib, conivaptan, brequinar, bedaquiline, celecoxib, gemcitabine, clofazimine, tolcapone with potential antiviral properties. **Sean Ekins** et al. [49, 50] provide a detailed explanation of the drug discovery method for Zika and Ebola viruses and further illustrate that there might be lacunas in the drug discovery process. They further explain the attempts of drug discovery for COVID-19 and whether existing antiviral therapeutics can be repurposed for SARS-CoV-2 virus. **Bo Ram Beck** et al. [51] have presented a DL model called Molecule Transformer-Drug Target Interaction (MT-DTI) to identify existing drugs that can be used against the SARS-CoV-2 virus. They have identified a list of existing antiviral drugs that can be considered for the treatment of COVID-19. According to the MT-DTI model, the atazanavir is considered as the best compound against SARS-CoV-2 3C like protease. They also propose that antiviral agents like Kaletra could be used as a treatment for COVID-19.

**3. Application of Deep Learning and AI for Covid-19 Detection and Screening**

The spread of COVID-19 can be limited by the use of Machine Learning (ML) and Deep Learning systems for early detection and screening of patients [52]. With the help of DL based-model, the COVID-19 pneumonia can be easily detected from CT, ultrasound, and Xray images [53, 54]. These systems will perform classification between COVID-19, normal, bacterial pneumonia and will also segment and quantify the infection sites [55]. In brief, the main tasks to be undertaken by a DL-based system are enlisted as follows:

1. Binary Classification to distinguish COVID-19 cases from healthy cases for proper detection and diagnosis.
2. Multi-class classification to correctly classify COVID-19 pneumonia, community-acquired pneumonia, bacterial pneumonia, and healthy cases.
3. Delineation of the affected regions in the lungs using the process of segmentation.
4. Quantification of the level of infection in the lungs using deep learning models.
5. Overall decision support for clinicians in the healthcare systems.

For the task of binary classification, the supervised DL models are provided with suitable datasets of normal and COVID-19 pneumonia images and class labels [56, 57]. These models extract deep features from the images and map them to their respective category's labels. The fully trained models are then used for the classification of new Xray or CT images that were not used in training. According to the survey conducted, most of the research works employ either of the two modalities namely the Xray images and CT scans for automatic detection using Deep Learning (DL) architectures [58, 59].

## 3.1 Data Collection and Pre-processing:

The first step in the development of an automatic COVID-19 diagnosis system is the collection of data that is required to train the DL models. In this scenario, the CT scans and X-Ray images form the most important modalities that can be used. The collected data is then pre-processed and augmented by operations like normalization, contrast enhancement, rotation, horizontal and vertical flip so that the CNN model can extract relevant features. **Joseph Paul Cohen** et al. [60] describe the publically available COVID-19 image database with 123 frontal view X-Ray images. The database is developed with the help of various open access repositories and publications as shown in fig 3. Various sources of CT and X-Ray images include the COVID-19 CT segmentation dataset, Lung Image Database Consortium (LIDC) dataset, COVID-CT, HUG dataset, COVID-chest X-ray dataset, Kaggle RSNA Pneumonia Detection Dataset, Chest X-Ray Images, and COVID-19 database [61].

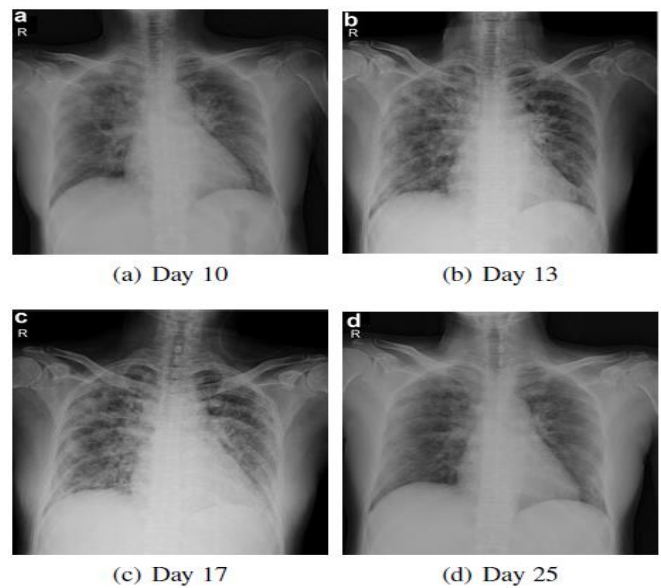

Fig. 3:Example X-Ray images from the same patient (#19) on different days extracted from Cheng et al. [60]

## 3.2 CNN and feature extraction:

The DL models perform end-to-end training by a sequence of operations like convolution and pooling on the input modality like CT or X-Ray image [62]. Furthermore, various types of activation functions like ReLU, sigmoid, and SoftMax functions are applied on the feature maps to make the system non-linear. With the help of these operations feature vectors are obtained which can then be utilized for tasks like classification, segmentation, and quantification of the disease as shown in fig 4. Many research works are aimed at employing the pre-trained models directly for COVID-19 diagnosis like AlexNet, Visual Geometry Group (VGG-16/19), U-Net, ResNet, Xception, and GoogLeNet. Another approach taken involves the development of new and improved DL-based architectures that are trained on the CT or Xray datasets available. For the diagnosis of COVID-19, many DL-based models have been utilized employing publically available datasets. The DL models are trained using three approaches namely supervised, unsupervised, and reinforcement learning.

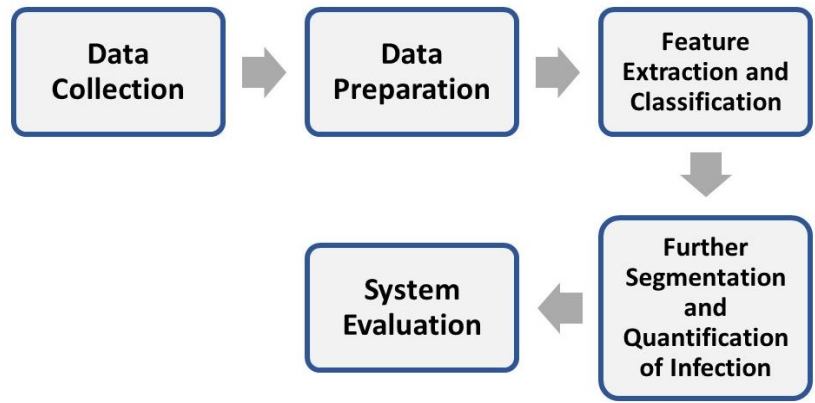

Fig 4:Sequence of Steps followed by a DL framework

### 3.3 Evaluation of Classification Task:

For evaluation of the Deep Learning models for the classification task, various parameters can be utilized. These evaluation parameters include confusion matrix, accuracy, specificity, recall, precision, etc [63]. The confusion matrix represents a matric of correct as well as false classifications as shown in Fig 5. The four values represented in the confusion matrix include True Positive (TP), True Negative (TN), False Positive (FP), and False Negative (FN). With the help of the values obtained from the confusion matrix, other evaluation parameters like the accuracy, specificity, recall, and precision can be further calculated [64]. The TP value specifies the number of cases that have been correctly identified as having the disease which in our case is the COVID-19 pneumonia. In other words, they have been correctly or truly classified as COVID-19 positive. The TN category includes cases that don't have the disease and have been correctly identified as normal or COVID-19 negative. The other two categories of FN and FP indicates the number of cases for which the model has misdiagnosed. FN indicates the number of cases that are falsely diagnosed as normal or COVID-19 negative but have the disease [65]. This can be very dangerous as the person will not know that he has the disease and will not get timely treatment or support and can also spread the virus leading to more cases. The other category of False Positive is falsely diagnosed to be COVID-19 positive or in general, the system predicts that they have the disease when they are normal. This again can lead to a lot of mental and physical disturbance to the person involved. So, any DL classification algorithm aims to increase the number of TP and TNs and reduce to zero the FP and FN cases.

|  | **Predicted Positive** | **Predicted Negative** |
|---|---|---|
| **Actual Positive** | True Positive (TP) | False Negative (FN) |
| **Actual Negative** | False Positive (FP) | True Negative (TN) |

Fig 5: Confusion Matrix

### 3.3.1 Accuracy:

The accuracy gives a percentage of how many cases have been correctly identified i.e. TP and TN from the total number of cases. This is shown in equation 1.

$$Accuracy = \frac{TN + TP}{(TN + TP + FN + FP)} \qquad (1)$$

### 3.3.2 Precision:

As we know that TP gives the number of cases that are correctly identified as having the disease and FP gives a false classification of COVID-19 positive who don't have the disease. TP+FP will provide the number of cases that the model has identified as having the disease out of which only TP values represent correct classification. The precision value gives the percentage of TP out of the total positive cases diagnosed by the system as shown in equation 2.

$$Precision = \frac{TP}{(TP + FP)} \qquad (2)$$

### 3.3.3 Recall:
It is also called as sensitivity and is represented as a ratio between the TP values and the summation of TP and FN as shown in equation 3. The FN identifies the falsely diagnosed cases as COVID-19 negative that in reality have the disease. The ratio will provide an insight into the number of cases that have been correctly diagnosed as COVID-19 positive from the total number of COVID-19 positive cases whether diagnosed correctly or incorrectly.

$$Recall = \frac{TP}{(TP + FN)} \qquad (3)$$

### 3.3.4 F1-Score:
The F1-Score also provides a measure of the classification models' accuracy in terms of Precision and Recall as shown in equation 4.

$$\text{F1 Score} = 2 \times \frac{(Precision \times Recall)}{(Precision + Recall)} \qquad (4)$$

### 3.3.5 Specificity:
Similarly, specificity provides a ratio between the correctly or truly diagnosed negative cases i.e. normal cases, and the summation of TN and FP cases that indicates the total number of normal cases seen by the model.

$$Recall = \frac{TN}{(TN + FP)} \qquad (5)$$

## 4. Role of AI for COVID-19 detection Using CT scans:

The DL models can be used for COVID-19 diagnosis, effective screening, case classification, segmentation, and quantification of the infection.

**Shuai Wang** et al. [66] propose the first method to effectively screen and diagnose COVID-19 patients from other viral pneumonia cases. Their method consists of three main steps: 1) Pre-processing; 2) Feature extraction of region of Interest (ROI) patches and training; 3) Classification of the CT scan into COVID-19 or other viral pneumonia categories. Their proposed model is built upon a pre-trained Inception network which extracts features from the input images to form feature vectors. The other part of the network includes a fully connected network for classification. They collected 1065 CT scans of COVID-19 cases along with other viral pneumonia patients. The internal validation achieved an accuracy of 89.5% with a sensitivity of 0.87, and specificity of 0.88. Furthermore, the external validation of their model yielded values of total accuracy, specificity, and sensitivity of 79.3%, 0.83, and 0.67 respectively.

**Xiaowei Xu** et al. [67] present a screening model to distinguish between COVID-19 pneumonia, Influenza-A viral pneumonia, and Normal cases. In their multi-center study, CT images were acquired from three COVID-19 designated hospitals in China. Out of the 618 CT images collected, 219 from 110 patients had COVID-19, 224 had Influenza-A viral pneumonia, H1N1, H3N2, H5N1, H7N9, etc, and 175 CT scans were obtained from healthy individuals. (85.4%) were used for training and validation sets and the remaining 14.6% were used as the test set. A 3D CNN model is used to segment out the pre-processed CT scan into multiple image patches. These patches are classified in the above mentioned 3 categories and confidence scores are generated. The Noisy-or Bayesian function is used to calculate the total confidence score and infection type for the entire CT scan with an accuracy of 86.7 %.

**Lin Li** et al. [68] have proposed a method to automatically detect the effect of COVID-19 using the Chest CT scans and simultaneously distinguish it from Community-acquired pneumonia (CAP). The dataset was collected from 6 hospitals with 4356 chest CT exams from 3,322 patients. They have developed a multi-center 3D deep learning model called COVNet with ResNet50 as the backbone. Initially, a 3D CT scan is pre-processed and the Region of Interest (ROI) is extracted using the U-net model and subsequently, the COVNet is utilized to detect COVID-19. The per exam sensitivity for the test set is 114 of 127 and specificity for detecting COVID-19 is 294 of 307 with an Area Under the Curve (AUC) of 0.96.

**Shuo Wang** et al. [69] present a fully automatic diagnostic DL network for COVID-19 detection and survival prediction or prognosis evaluation using CT scans. They collected 5372 CT scans from 7 hospitals from 2 main datasets namely the COVID-19 dataset with 1266 scans and the CT-epidermal growth factor receptor (CT-EGFR) dataset with 4106 scans. A total of 471 cases with COVID-19 are utilized for prognostic analysis and regular follow-up for at least 5 days was undertaken. For the training set, the diagnostic performance in terms of AUC, sensitivity, and specificity is equal to 0.90, 78.93%, and 89.93% respectively. Two other external validation sets are formulated and the performance of the model is evaluated with AUC = 0.87 and 0.88, sensitivity = 80.39% and 79.35%, specificity = 76.61% and 81.16%.

**Xuehai He** et al. [70] developed a Self-Trans approach which integrates self-supervised learning with transfer learning to achieve improved diagnostic accuracy for small datasets. They further develop a dataset containing hundreds of publically-available positive COVID-19 CT scans. Their model approaches an AUC of 0.94, an F1 score of 0.85 for detecting COVID-19 when trained with a few hundred CT scans. **Shaoping Hu** et al. [71] have proposed a weakly supervised deep learning method for diagnosis and classification of COVID-19. They perform binary classification of COVID-19 from non-COVID-19 cases and can detect infectious areas in CT images with high accuracy. They perform three-way classification and binary classification with two learning configurations called single-scale learning and multi-scale learning. The different configurations are compared in terms of accuracy, precision, sensitivity, specificity, and AUC.

**Hengyuan Kang** et al. [72] develop a system to extract features from CT images and automatically diagnose COVID-19 pneumonia and distinguish it from CAP. A unified latent representation is learned to fully encode different aspects of the features. They utilized a total of 2522 CT images among which 1495 cases are from COVID-19 patients and 1027 cases are from CAP patients. With the proposed method diagnosis performance is improved with evaluation parameters like accuracy, sensitivity, and specificity of 95.5%, 96.6%, and 93.2% respectively. **Deng-Ping Fan** et al. [73] present a novel COVID-19 Deep Network called Inf-Net for lung Infection Segmentation is proposed to automatically identify infected regions from chest CT images. **Vruddhi Shah** et al. [74] have developed a CTnet-10 deep learning model for COVID-19 diagnosis with 82.1 % accuracy. They also evaluated pre-trained models for binary classification of COVID-19 cases from non-COVID-19 cases like the DenseNet-169, ResNet-50, VGG-16, VGG-19, and InceptionV3. According to their evaluation VGG-19 performed the best with an accuracy of 94.52%. They also evaluated that DenseNet-169 has the second-highest accuracy of 93.15 %.

**Ali Abbasian Ardakani** et al. [75] evaluated the performance of 10 pre-trained CNNs like the VGG-16, AlexNet, VGG-19, GoogleNet, ResNet-18, ResNet-50, ResNet-101, SqueezeNet, Xception and MobileNet-V2. They utilized 1020 COVID-19 CT images and 86 cases having non-COVID atypical and viral pneumonia. They evaluated that the ResNet-101 achieved the best performance with an AUC of 0.994, sensitivity equal to 98.04% and specificity of 100%, and accuracy equal to 99.51%. Xception model also achieved comparable results of AUC, sensitivity, specificity, and accuracy equal to 0.994, 98.04%, 100%, and 99.51%.

**Zhongyi Han** et al. [76] propose a model called attention-based deep 3D multiple instance learning (AD3D-MIL) in which labels are assigned to CT scans which are taken as a bag of instances. The Bernoulli distribution is calculated for the bag-level labels for accurate screening of COVID-19 patients. Their proposed algorithm achieves an accuracy of 97.9% with an AUC of 99.0%, and Cohen Kappa score of 95.7% with weak labels. They have compared their results with earlier works of D. Tran et al. and Chuansheng Zheng et al. as shown in Tables 2 and 3.

**Chuansheng Zheng** et al. [77] report a weakly supervised DL model called DeCoVNet for the detection of COVID-19 with weak labels using the 3D CT scans. The U-Net model is used to first segment the lung region that is given into a 3D Deep Neural Network to calculate the probability of occurrence of COVID-19. A total of 499 CT volumes was used for training the model and 131 CT volumes for testing purposes.

Table 2: Classification results for Binary classification between COVID-19 and NON-COVID-19 [76]

| Metric | Method | | |
|---|---|---|---|
| | **D. Tran et al. [78]** | **C. Zheng et al. [77]** | **AD3D-MIL [76]** |
| Accuracy | 96.8 | 96.8 | 97.9 |
| AUC | 98.2 | 98.2 | 99.0 |
| F1 Score | 96.8 | 96.8 | 97.9 |
| Precision | 96.8 | 96.8 | 97.9 |
| Recall | 96.8 | 96.8 | 97.9 |
| Cohen Kappa Score | 93.6 | 93.6 | 95.7 |

Table 3: Classification with 3 classes: 1) COVID-19; 2) Common pneumonia; 3) Normal [76]

| Metric | Method | | |
|---|---|---|---|
| | **D. Tran et al. [78]** | **C. Zheng et al. [77]** | **AD3D-MIL [76]** |
| Accuracy | 89.7±0.9 | 90.6±0.6 | 94.3±0.7 |
| AUC | 97.1±0.4 | 97.5±0.1 | 98.8±0.2 |
| F1 Score | 86.1±0.5 | 86.1±0.3 | 92.3±0.4 |
| Precision | 88.2±0.2 | 93.7±0.5 | 95.9±0.3 |
| Recall | 85.0±0.3 | 84.1±0.6 | 90.5±0.5 |
| Cohen Kappa Score | 83.7±0.8 | 84.9±0.7 | 91.1±0.9 |

**Fei Shan** et al. [79] presented a model to automatically segment and quantify the lung infection volume in CT images using Deep Learning. In total 300 COVID-19 CT images were collected from Shanghai Public Health Clinical Center for validation and 249 COVID-19 CT images were obtained from other centers that were outside Shanghai. The multi-center system employed "VB-Net" neural network to segment COVID-19 infection regions. The training was implemented using the Human-in-the-Loop (HITL) strategy with 249 COVID-19 cases, and validated using 300 new COVID-19 cases. The system is evaluated using various parameters like the Dice Coefficient (DC) equal to 91.6% ± 10.0%, Percentage of Infection (POI), and difference of volume is calculated between the manual delineation and results obtained by the model. **Longxi Zhou** et al. [80] have developed a multi-center model to segment and quantify the infection regions from CT scans. They collected 201 CT scans of 140 COVID-19 patients from 4 different hospitals in Heilongjiang Province, China, and 20 COVID-19 CT scans from King Faisal Specialist Hospital and Research Center (KFSHRC) in Riyadh, Saudi Arabia. They have presented a fully automatic and machine agnostic method including the development of a CT simulator to incorporate all the changes in patient's data collected at different times. Furthermore, they present a novel deep learning algorithm that decomposes a 3D segmentation problem into 3 2D problems thus reducing complexity. The segmentation performance on the Harbin dataset is evaluated using Dice Coefficient (DC), recall, and worst-case with values equal to 0.783±0.080, 0.776±0.072, and (0.577, 0) respectively. They can reduce the problem of data scarcity and further can identify small objects in a bigger scenario. **Hai-tao Zhang** et al. [81] present a model to diagnose, localize, and quantify the COVID-19 infection. A total of 2460 CT scans were obtained from Huoshenshan Hospital in Wuhan, China, and a uAI Intelligent Assistant Analysis System was used to assess the images. The deep learning-based software is composed of a modified 3D CNN and a V-Net that is used to analyze the infectious regions and calculate the distribution and volume of infection in the lungs. Out of the 2460 CT images, 2215 (90%) had clinical manifestations of COVID-19 pneumonia in both lungs. Table 4 presents a comparison of the techniques discussed above in terms of accuracy.

Table 4: Comparison of schemes based on the highest accuracy achieved for detection using CT images.

| Proposed Model | Highest Achieved Accuracy (%) in Classification and Detection |
|---|---|
| Shuai Wang et. al. [66] | 79.3%, |
| Xiaowei Xu et al. [67] | 86.7 % |
| Hengyuan Kang et al [72] | 95.5%, |
| Vruddhi Shah et al. [74] (VGG19) | 94.52%. |
| Ali Abbasian Ardakani et al. [75] (ResNet101) | 99.51%. |
| Zhongyi Han et al. [76] | 97.9% |

## 5. Role of AI for COVID-19 detection Using Xray Images:

Many efforts have been made to detect COVID-19 with DL models most of which are built on pre-trained models. The pre-trained models have been already trained on enormous databases of images like the ImageNet dataset and can be repurposed for COVID-19 detection [82]. This is known as Transfer learning and can be extremely useful for application wherein data samples are insufficient. Several research works including pre-trained models used with Transfer Learning and Custom models are discussed below.

**Tulin Ozturk** et al. [83] present a new DL model for detection, binary, and multi-class classification using X-ray images. They selected the Darknet-19 classifier model that forms the underlying framework for the real-time object detection model called You-Only-Look-Once (YOLO). Inspired by the Darknet architecture, they have designed the DarkCovidNet architecture which can detect as well as classify between COVID-19 vs normal vs pneumonia categories. The model can achieve an accuracy of 98.08% for binary classification and 87.02% for multi-class classification.

**Nour Eldeen M. Khalifa** et al. [84] propose a DL model based on Generative Adversarial Networks (GANs) and transfer learning for detection of COVID-19 using a small Chest X-ray dataset. The GAN is used to augment the dataset and prevents the model from overfitting. Although the dataset had 5863 images but only 10% of those were used for training and the rest were developed using the GAN and had only two categories: Normal and Pneumonia. They selected the pre-trained models like GoogLeNet, AlexNet, Squeeznet, and Resnet18 and further utilized transfer learning to perform the binary classification. They evaluated that Resnet18 has the most superior performance with a testing accuracy of 99.00%.

**Pedro R. A. S. Bassi and Romis Attux** [85] propose a DL model to detect and distinguish COVID-19 cases from viral pneumonia and normal cases. They fine-tuned models already trained on ImageNet and applied the transfer learning approach twice to train the model on the NIH ChestX-Ray14 dataset. Furthermore, they adopted strategies like output neuron keeping and used Layer-wise Relevance Propagation (LRP) to produce heatmaps. The model can achieve an accuracy of 99.4% and an F1 score of 0.994.

**Ali Narin** et al. [86] present a CNN-based model for automatic detection of COVID-19 using chest X-ray radiographs. They have evaluated three different CNN-based models like the InceptionV3, ResNet50 and, Inception-ResNetV2 for their applicability in the COVID-19 detection task. The pre-trained ResNet50 model achieves the highest classification accuracy of 98%. The other two models: InceptionV3 and Inception-ResNetV2 achieve accuracies of 97% and 87% respectively. The dataset used by them includes 224×224 sized 100 images in total with 50 having COVID-19 and 50 belonging to the normal category.

**Ezz El-Din Hemdan** et al. [87] have developed COVIDX-Net DL-classifier for COVID-19 detection using X-ray images. Their proposed model using pre-processing steps on the input data like scaling and one-hot encoding. Subsequently, seven different architectures of CNNs like the VGG19 and Xception, DenseNet201, and MobileNetV2 among others are utilized for accurate classification. The dataset used in this model contains 50 X-ray images that are split into training, validation, and testing set. They conclude that the VGG19 and DenseNet201 models perform the best for classification with accuracy up to 90% and InceptionV3 is identified to have the worst performance with an accuracy of 50%.

**Ioannis D. Apostolopoulos** et al. [88] present a transfer learning-based approach for the automatic detection of COVID-19 cases. They perform multi-class classification with 3 categories: 1) Common pneumonia, 2) COVID-19, and 3) Normal Category. The number of images in the three categories include 224 for COVID-19, 700 for Common pneumonia, and 504 for the normal category with a total number equal to 1427. They conclude that the VGG19 and MobileNet can achieve the best accuracy among the various models that also include Inception, Xception, and Inception ResNetV2. The overall accuracy of the method reaches up to 97.8% as shown in Table 5.

Table 5: Comparison of different models on the dataset mentioned in [88]

| Network | Accuracy 2-Class (%) | Accuracy 3-Class (%) | Sensitivity (%) | Specificity (%) |
|---|---|---|---|---|
| **VGG19 [89]** | 98.75 | 93.48 | 92.85 | 98.75 |
| **MobileNet [90]** | 97.40 | 92.85 | 99.10 | 97.09 |
| **Inception [91]** | 86.13 | 92.85 | 12.94 | 99.70 |
| **Xception [92]** | 85.57 | 92.85 | 0.08 | 99.99 |
| **InceptionResNetV2 [91]** | 84.38 | 92.85 | 0.01 | 99.83 |

**Asif Iqbal Khan** et al. [93] have proposed a DL model called CoreNet which can detect COVID-19 infection from X-Ray images. They have utilized an Xception architecture pre-trained on the ImageNet dataset and then perform end-to-end training on the publically available Chest-Xray images. They perform binary, 3-class, and 4-class classification with categories being COVID-19, Bacterial Pneumonia, Viral Pneumonia, and Normal class. The overall accuracy of the CoroNet model is 89.6% with precision and recall for COVID-19 are equal to 93% and 98.2% for 4-class classification. For the 3-class classification with categories: Normal vs COVID-19 vs Pneumonia, the model can achieve an accuracy of 95%.

**Sohaib Asif** et al. [94] present a Deep CNN (DCNN)-based model for automatic detection of COVID-19 pneumonia using Chest X-Ray images. They utilize the InceptionV3 with transfer learning to maximize accuracy. The dataset contains X-Ray images with 1341 for Normal category, 864 for COVID-19, and 1345 for viral pneumonia. The accuracy of the system is more than 98% with a training accuracy of 97% and a validation accuracy of 93%.

**Prabira Kumar Sethy** et al. [95] present a scheme to classify the Xray scans of healthy patients, other Pneumonia cases (bacterial and viral), and COVID-19 pneumonia. They evaluate 13 different pre-trained CNNs for feature extraction and utilize the Support Vector Machine (SVM) for final classification. The ResNet50 model with SVM achieves an accuracy of 98.66%, sensitivity of 95.33%. The comparison of the schemes is presented in Table 6.

Table 6: Comparison between schemes utilizing X-Ray images

| Proposed Scheme | Model and Underlying Architecture | Dataset (X-Ray) | Accuracy (%) for Classification |
|---|---|---|---|
| Tulin Ozturk et al. [83] | DarkNet-19 | ChestX-ray8 database and image database by Cohen JP | 98.08 |
| Nour Eldeen M. Khalifa et al. [84] | GoogLeNet, AlexNet, Squeeznet, and Resnet18 | dataset had 5863 (used only 10%) + GAN augmentation | 99.00 (ResNet18) |
| Ali Narin et al. [86] | InceptionV3, ResNet50 and, Inception-ResNetV2 | 224×224 sized 100 images. (50 for COVID-19 and 50 for Normal) | 98 (ResNet50) |
| Ezz El-Din Hemdan et al. [87] | VGG19, Xception, DenseNet201, InceptionV3, ResNetV2, InceptionResNetV2, and MobileNetV2 | 50 images | 90 (VGG19 and DenseNet201) |
| Ioannis D. Apostolopoulos et al. [88] | VGG19, MobileNet, Inception, Xception, InceptionResNetV2 | 1427 images (224 for COVID-19, 700 for Common pneumonia, and 504 for the normal category) | 97.8 (VGG19 and MobileNet) |
| Asif Iqbal Khan et al. [93] (CoroNet) | CoroNet, Xception | 1251 images (310 for Normal, 330 for Bacterial Pneumonia, 327 for | 89.6 (4-class) 95 (3-class) 99 (binary) |

|  |  | Viral Pneumonia, and 284 for COVID-19) | Overall accuracy of 90% for the dataset prepared by Ozturk et al. [83] |
|---|---|---|---|
| Sohaib Asif et al. [94] | InceptionV3 | dataset contains X-Ray images with 1341 for Normal category, 864 for COVID-19, and 1345 for viral pneumonia | 98% |

**Linda Wang** et al. [96] have developed the publically available COVID-Net model and COVIDx chest X-ray dataset to detect COVID-19. The COVIDx dataset contains 13,975 X-ray images from 13870 patients with a combination of images from 5 publically available datasets including the COVID-19 Image Data Collection and COVID-19 radiography database among others. The COVID-Net model is pre-trained on the ImageNet dataset and subsequently trained on the COVIDx dataset. The COVID-Net model has improved performance as compared to VGG-19 and ResNet-50 in terms of the number of parameters, number of multiply-accumulation (MAC) operations, and accuracy as shown in table 7. The model was able to achieve a high sensitivity of 91.0% and Positive Predicted value equal to 98.9% for COVID-19 cases.

Table 7: Comparison of COVID-Net with state-of-the-art models like VGG-19 AND ResNet50 [96]

| Architecture | Parameters (M) | MACs (G) | Accuracy (%) |
|---|---|---|---|
| VGG-19 [89] | 20.37 | 89.63 | 83.0 |
| ResNet-50 [97] | 24.97 | 17.75 | 90.6 |
| **COVID-Net [96]** | **11.75** | **7.50** | **93.3** |

**Harsh Panwar** et al. [98] present a new DL-based model called nCOVnet which can rapidly detect COVID-19 using X-Ray images. They apply pre-processing on the original dataset like rotation, horizontal and vertical flip, and rescaling of image size to $224 \times 224 \times 3$. The nCOVnet model has 24 layers including input and output with the remaining being a combination of Convolution, ReLU, and Max Pooling layers. They have taken into consideration the problem of data leakage and produce unbiased results. The model uses VGG-16 pre-trained on the ImageNet dataset. The overall accuracy of the model is 88.10% with sensitivity equal to 97.62% and specificity equal to 78.57%.

**Rachna Sethi** et al [99] propose a method to investigate four different CNN architectures for the diagnosis of COVID-19 using Chest X-Ray images. The models include Inception V3, ResNet50, MobileNet, and Xception all of which are pre-trained on the ImageNet dataset. They found the accuracies of the Inception V3, ResNet50, MobileNet, and

Xception to be 0.981, 0.825, 0.986, and 0.974. Accordingly, they conclude that the MobileNet has achieved the best performance out of the four evaluated models.

**Tayyip Ozcan** [100] report a method to detect COVID-19 using the X-Ray images with the help of Grid Search (GS) and pre-trained CNN. They evaluate three different models: 1) GoogleNet; 2) ResNet18; and 3) ResNet50 for classification between 4 categories like Bacterial pneumonia, Viral pneumonia, Normal cases, and COVID-19. The combination of GS with ResNet50 yields the best performance equal to 97.69%.

**Kabid Hassan Shibly** et al. [101] present a Faster R-CNN framework to detect COVID-19 in patients using the X-Ray images. Their proposed method can achieve a binary classification accuracy of 97.36%, sensitivity and, precision of 97.65%, and 99.28% respectively. The custom dataset utilized by them consists of 5450 X-Ray images from the COVID Chest X-Ray dataset developed by Cohen et al. and the RSNA pneumonia dataset. They also used the COVIDx dataset with a total of 13800 X-Ray images.

Table 8:Comparison of Different Models using X-Ray and CT image Modalities [101]

| Modality | Method | Accuracy (%) |
|---|---|---|
| X-Ray | COVID-Net [L. Wang et al.] | 92.40 |
| X-Ray | ResNet50+ SVM [P. K. Sethy et al.] | 95.38 |
| X-Ray | Deep CNN ResNet-50 [A. Narin et al.] | 98.00 |
| X-Ray | VGG-19 [I. D. Apostolopoulos et al.] | 93.48 |
| X-Ray | COVIDX-Net [E. E. D. Hemdan] | 90.00 |
| CT Scan | M-Inception [S. Wang et al.] | 82.90 |
| CT Scan | ResNet [X. Xu et al.] | 86.70 |
| X-Ray | DarkCovidNet [T. Ozturk et al.] | 87.02 |
| X-Ray | Faster R-CNN [K. H. Shibly et al.] | 97.36 |

## 6. Conclusion

COVID-19 pandemic is an overwhelming challenge for the medical community around the globe. To tackle the effects of the outbreak AI and DL-based methods can be employed in the areas like screening, treatment, vaccine development, contact tracing and, survival prediction. In this study, we have presented an overview of the different application areas of DL and further explained in detail its use in the sub-task of detection and diagnosis of COVID-19 with the help of imaging modalities like CT-scan and X-Ray images. Several methods have been proposed that employ CNN-based models either custom-built or developed over a pre-trained model like VGG-19, ResNet50 among others. These models are not only able to detect and diagnose COVID-19 but can also classify COVID-19 pneumonia from CAP, other viral pneumonia, and the normal cases with a high level of accuracy. Furthermore, the tasks of segmentation and infection quantification can also be effectively performed with the help of the DL models. It can be concluded that the DL-based systems can assist the healthcare professional in developed as well as rural areas to mitigate the challenge of COVID-19.